\documentclass[10pt,twocolumn,letterpaper]{article}

\usepackage{iccv}
\usepackage{times}
\usepackage{epsfig}
\usepackage{graphicx}
\usepackage{amsmath}
\usepackage{amssymb}
\usepackage{multirow}
\usepackage{pifont}
\newcommand{\cmark}{\ding{51}}
\newcommand{\xmark}{\ding{55}}

\usepackage[pagebackref=true,breaklinks=true,letterpaper=true,colorlinks,bookmarks=false]{hyperref}

\iccvfinalcopy % *** Uncomment this line for the final submission

\ificcvfinal\fi
\begin{document}

%%%%%%%%% TITLE
\title{Neural Deformable Models for 3D Bi-Ventricular Heart Shape Reconstruction and Modeling from 2D Sparse Cardiac Magnetic Resonance Imaging}

\author{Meng Ye$^{1}$, Dong Yang$^{2}$,  Mikael Kanski$^{3}$,
Leon Axel$^{3}$, Dimitris Metaxas$^{1}$\\
$_{}^{1}\textrm{}$Rutgers University, $_{}^{2}\textrm{}$NVIDIA, $_{}^{3}\textrm{}$New York University School of Medicine\\
{\tt\small \{my389, dnm\}@cs.rutgers.edu }
}

\maketitle
% Remove page # from the first page of camera-ready.
%\ificcvfinal\thispagestyle{empty}\fi

%%%%%%%%% ABSTRACT
\begin{abstract}
We propose a novel neural deformable model (NDM) targeting at the reconstruction and modeling of 3D bi-ventricular shape of the heart from 2D sparse cardiac magnetic resonance (CMR) imaging data. 
We model the bi-ventricular shape using blended deformable superquadrics, which are parameterized by a set of geometric parameter functions and are capable of deforming globally and locally. 
While global geometric parameter functions and deformations capture gross shape features from visual data, local deformations, parameterized as neural diffeomorphic point flows, can be learned to recover the detailed heart shape.
Different from iterative optimization methods used in conventional deformable model formulations, NDMs can be trained to learn such geometric parameter functions, global and local deformations from a shape distribution manifold. 
Our NDM can learn to densify a sparse cardiac point cloud with arbitrary scales and generate high-quality triangular meshes automatically. 
It also enables the implicit learning of dense correspondences among different heart shape instances for accurate cardiac shape registration.
Furthermore, the parameters of NDM are intuitive, and can be used by a physician without sophisticated post-processing.
Experimental results on a large CMR dataset demonstrate the improved performance of NDM over conventional methods.
Project page is at \url{https://github.com/DeepTag/NeuralDeformableModels}.
\end{abstract}

%%%%%%%%% BODY TEXT
\section{Introduction}
\label{sec:intro}

\begin{figure*}[!t]
\begin{center}
\includegraphics[width=1.0\linewidth]{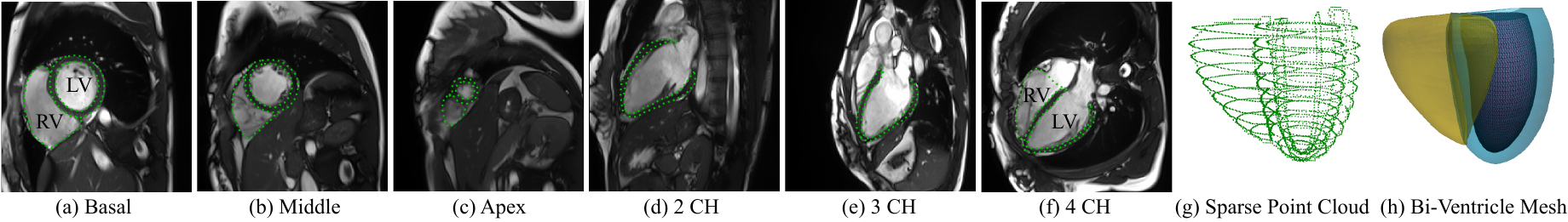}
\end{center}
   \caption{Cardiac MR standard scan views and bi-ventricular shape reconstruction. (a-c) Short axis (SAX) views at the basal/middle/apex region of the left ventricle (LV). (d-f) Long axis (LAX) 2/3/4 chamber (CH) views. Green points show the endo-/epi-cardial border of LV and the border of right ventricle (RV). (g) Sparse cardiac point cloud with 10 SAX and 3 LAX slices' data. (h) Bi-ventricular mesh reconstructed from (g) by our neural deformable model, in which the LV endo-cardial, epi-cardial and RV surfaces are shown in red, blue and yellow, respectively.}
\label{fig1}
\end{figure*}

Cardiac magnetic resonance imaging (CMR) is the gold standard for non-invasive evaluation of global cardiac function, i.e., blood pumping of the left ventricle (LV). Due to the relatively slow imaging speed of MR, current clinical CMR scan protocols are 2D-based. To recover the 3D geometry of the LV and right ventricle (RV), we usually scan a stack of short axis (SAX) and several long axis (LAX) images. As shown in Fig.~\ref{fig1} (a-f), SAX images cover the range from the base to the apex of the heart and LAX images include 2-, 3-, 4-chamber views. Although in-plane resolution of the image is high enough to capture 2D cardiac anatomical details, the through-plane resolution is much lower, in order to reduce imaging time. Therefore, the SAX and LAX images together can only produce a sparse 3D point cloud of the heart, as shown in Fig.~\ref{fig1} (g). The dense and accurate 3D geometry reconstruction, as shown in Fig.~\ref{fig1} (h), is thus needed for not only the downstream image analysis tasks, such as the estimation of LV mass and volume\cite{young2000left},  and 3D cardiac wall strain fields~\cite{park1996analysis, haber2000three, hu2003vivo}, but also other clinical applications, such as image-guided interventions~\cite{suinesiaputra2017statistical} and biomechanics finite element-based simulations~\cite{wang2013structure, ukwatta2015image, gurev2011models}.  

3D cardiac geometry reconstruction has a long history~\cite{nielsen1991mathematical, park1996analysis, young2000left, luo2005lv, van2006spasm, chen2021shape, romaszko2021neural, joyce2022rapid}. Most of the previous methods reconstruct the 3D shape from a sparse point cloud generated from segmentation results on CMR images. Those segmentations delineate the myocardium wall, blood pool of the LV and RV, as shown by the green points in Fig.~\ref{fig1}. 
As one of the conventional methods, template mesh adaptation first constructs a template mesh that describes the mean shape of the target and then registers this template with the sparse point cloud~\cite{van2006spasm}. The resulting meshes are usually not accurate for data that have large shape variation from the mean shape, which leads to a generalization problem of this kind of approach. 

We are inspired by the success of conventional deformable models (CDMs) for their efficiency in shape reconstruction and ability to provide intuitive and explainable shape parameters~\cite{park1996analysis, park2002lv, luo2005lv}. CDMs can efficiently describe the target shape using a set of global and local deformation parameters. However, to fit a shape primitive to a target shape, conventional methods use iterative optimization and fit the primitive to the given data. We propose a neural deformable model (NDM), which is composed of global deformations using parameter functions and local deformations using neural diffeomorphic flows. It can be trained to learn the global and local deformation parameters, conditioning on the sparse given data, from a shape distribution manifold.
As shown in Fig.~\ref{fig1} (h), NDM can accurately reconstruct the bi-ventricle shape from the sparse point cloud, as shown in Fig.~\ref{fig1} (g). NDM also enables implicit learning of dense correspondences among shape instances, which facilitates accurate shape registration. Lastly, NDM estimates shape parameters of the heart directly, which can be used by clinicians without sophisticated post-processing.

Our key contributions in this work can be summarized as follows:
(1) We formulate a new shape modeling framework which can capture global shape information and recover complex shape details accurately;
(2) We design a novel neural deformable model and propose an efficient coarse-to-fine learning paradigm to learn the global and local deformation parameters from a shape distribution manifold; 
(3) We show how such a shape modeling framework can be trained to perform clinically meaningful 3D heart shape reconstruction, registration and interpretation. Furthermore, our method has the potentials to inspire shape reconstruction and registration tasks in other domains.

%-------------------------------------------------------------------------
\section{Related Work}
\subsection{3D Heart Ventricle Reconstruction}
Most existing 3D heart ventricle shape reconstruction methods are based on template mesh registration. They aim to register a template mesh which has a dense set of vertices to a sparse input point cloud. 
Early works~\cite{nielsen1991mathematical, park1996analysis, young2000left, luo2005lv, van2006spasm} use finite element shape models to take advantage of basis function interpolation and extrapolation and perform a least square fitting procedure to fit the initial mesh to the sparse visual data. The fitting procedure is a time-consuming iterative process. 
A recent learning-based template mesh registration work~\cite{chen2021shape} can learn to reconstruct bi-ventricular shape from ground truth shape data, enabling the usage of not only observed sparse visual data but also a shape distribution manifold. However, the low resolution of its 3D binary volume, used as the bridge between the input and template for shape feature transferring, decreases the 3D shape reconstruction accuracy. 
The main drawback of the template mesh registration method is the lack of generalization, when the shape variation between the template and the target is too large to accurately register them. 
Our method decomposes the geometry correspondence between the shape primitive and the target into a set of global and local deformation parameters, which are learned in a coarse-to-fine fashion, resulting in accurate shape reconstructions even in the presence of large variations between the shape primitive and the target. 

% \subsection{Point Cloud Upsampling}
% Previous methods for point cloud upsampling could be mainly categorized as optimization- and deep learning-based. The optimization-based methods~\cite{alexa2003computing, lipman2007parameterization, huang2009consolidation} rely on shape priors, such as the surface smoothness and normal distributions. Apart from the lack of generalization to diverse 3D shapes, these methods are iteratively optimized, which is time-consuming.
% Deep learning-based methods rely on deep point feature extraction which becomes possible after the introduction of PointNet~\cite{qi2017pointnet} and PointNet++~\cite{qi2017pointnet++}. They generate new points via multi-branch MLPs~\cite{yu2018pu, yu2018ec} or point feature duplication~\cite{yifan2019patch, li2019pu, li2021point}. Some work makes use of structural 3D spatial neighborhood information in which they build GCNs~\cite{qian2021pu} or edge vectors~\cite{luo2021pu} for new point generation.
% More recently, neural fields~\cite{feng2022neural} and neural representation~\cite{zhao2022self} are introduced for point cloud upsampling. 
% All of these methods only learn point relationship between the sparse input and the dense target and cannot generate a mesh automatically.
% Different from all of these methods, our method exploits the implicitly learned correspondence between the shape primitive and the target, to upsample the point cloud and recover the mesh simultaneously. 

\subsection{Deformable Models}
Deformable models~\cite{kass1988snakes, terzopoulos1988deformable, terzopoulos1988constraints, terzopoulos1991dynamic, decarlo1998shape} are in the form of 2D curves,  3D surfaces or volumes whose points are parameterized in the material coordinate domain. Such models, also called deformable primitives, are deformed under the effect of external and internal forces. External forces are computed from the visual data, to drive the model to fit the target data. Internal forces are defined within the model itself, to preserve its smoothness during deformation.  
%There are two kinds of deformable models, i.e., parametric deformable models~\cite{kass1988snakes, terzopoulos1991dynamic} based on explicit parametric forms and geometric deformable models~\cite{malladi1995shape, caselles1997geodesic} based on curve evolution~\cite{sapiro1993affine} and level set~\cite{osher1988fronts}. Compared with geometric deformable models that need post-processing of the level set for shape parameterization, parametric deformable models enables direct interaction with the model and can be incorporated with prior target shape information. 
In~\cite{park1996deformable, park1996analysis}, constant geometric parameters of a primitive are extended to parameter functions to capture local shape variations. In our work, due to the aforementioned advantage, we use this parameter function-based primitive representation and combine it with a neural diffeomorphic flow~\cite{niemeyer2019occupancy, gupta2020neural, sun2022topology} for local deformations, to further improve shape reconstruction accuracy. 

\subsection{Diffeomorphic Flow}
Diffeomorphic flow is a spatial transformation that is smooth and whose inverse is smooth. In computational anatomy modeling of the brain, large deformation diffeomorphic metric mapping (LDDMM) has been used to compute a time-dependent velocity vector field based on an ordinary differential equation (ODE)~\cite{beg2005computing}. To overcome the computation complexity of LDDMM,
a stationary velocity field (SVF) is introduced to parameterize diffeomorphisms~\cite{arsigny2006log}. Recent work on neural ordinary differential equations (NODE)~\cite{chen2018neural, chen2020learning} enables the solution of neural diffeomorphic flow~\cite{niemeyer2019occupancy, gupta2020neural, sun2022topology}. 
We propose to integrate parameter function-based primitive and neural diffeomorphic flow parameterized local deformation as a novel neural deformable model, which enables the efficient coarse-to-fine learning of its geometric and deformation parameters from a shape distribution manifold.

\begin{figure}[t]
\begin{center}
\includegraphics[width=0.95\linewidth]{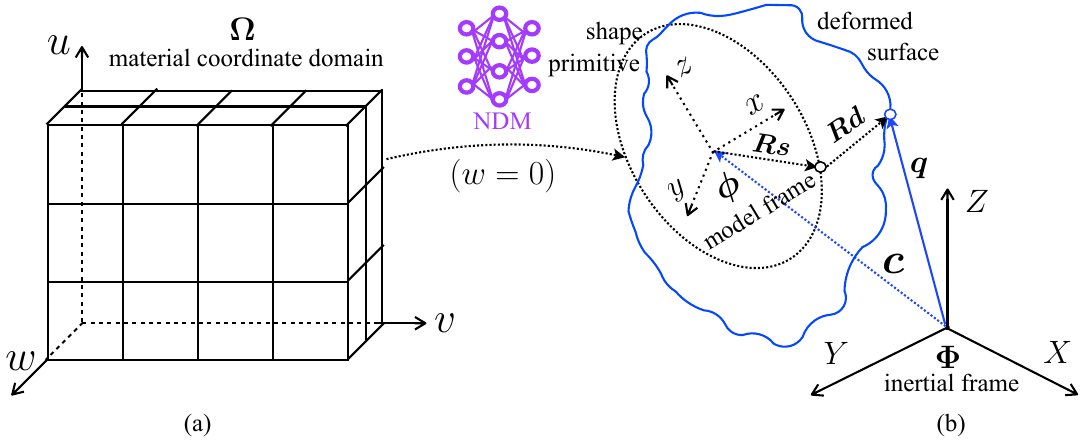}
\end{center}
   \caption{Geometry representation of a deformable surface model. (a) Material coordinate domain $\mathbf{\Omega}$. (b) The shape primitive and deformed surface mapped from the material coordinate domain (illustrated with $w=0$) by a neural deformable model (NDM).}
\label{fig2}
\end{figure}

\section{Method}
In this section, we introduce a neural deformable model (NDM) for bi-ventricle shape reconstruction, mesh generation and shape registration. We first provide the formulation of blended deformable superquadrics and then introduce a learning algorithm that can be applied to the NDM.

\subsection{General Geometry and Parameter Functions-based Deformable Models}
As shown in Fig.~\ref{fig2}, we formulate a 3D object surface model in space using material coordinates $\boldsymbol{m}=\left ( u, v, w \right )$, which are defined on a domain $\boldsymbol{\Omega }$. Given an inertial frame of reference $\boldsymbol{\Phi }$ in 3D, the positions of points $\boldsymbol{q}$ on the surface model relative to $\boldsymbol{\Phi }$ are represented by the following vector-valued function of $\boldsymbol{m}$:
\begin{equation}
    \boldsymbol{q}(\boldsymbol{m})=\left ( x(\boldsymbol{m}), y(\boldsymbol{m}), z(\boldsymbol{m}) \right )^{\top},
\end{equation}
where $\top$ is the transpose operator. To model the object pose, we introduce a model-centered reference frame $\boldsymbol{\phi }$ and represent the surface points positions as
\begin{equation}    \boldsymbol{q}=\boldsymbol{c}+\boldsymbol{R}(\boldsymbol{s} +\boldsymbol{d}),
\label{eq_qcsd}
\end{equation}
where $\boldsymbol{c}$ is the origin of model frame $\boldsymbol{\phi }$, $\boldsymbol{R}$ is the rotation matrix\footnote{We use a quaternion with unit magnitude to represent it.} that describes the orientation of $\boldsymbol{\phi }$, $\boldsymbol{s}$ is a shape primitive defined in the model frame $\boldsymbol{\phi }$, and $\boldsymbol{d}$ is a local deformation function.

\begin{figure}[t]
\begin{center}
\includegraphics[width=1.0\linewidth]{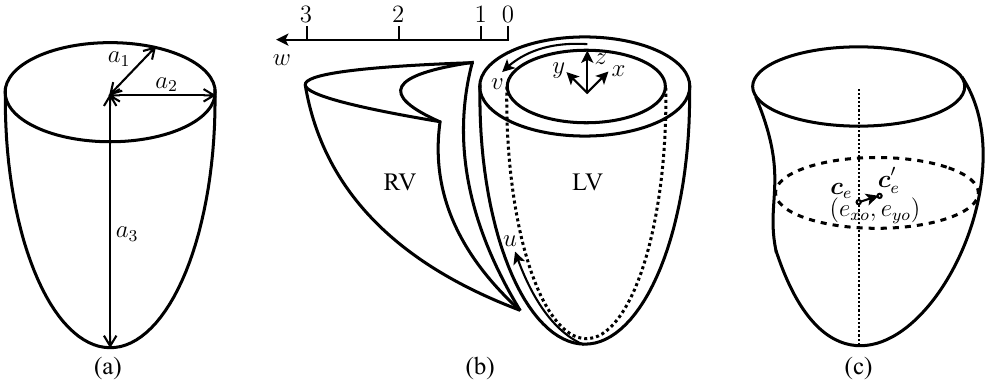}
\end{center}
   \caption{(a) A deformable shape primitive with aspect ratios $a_{1}$, $a_{2}$, $a_{3}$. (b) Blended deformable shape primitives for the modeling of bi-ventricular geometry. (c) Axis offset deformations $e_{xo}$, $e_{yo}$ that translate the center of the ellipse from $\boldsymbol{c}_{e}$ to $\boldsymbol{c}'_{e}$. }
\label{fig3}
\end{figure}

A shape primitive $\boldsymbol{s}$ could be a generalized cylinder~\cite{fang1994extruded}, geon~\cite{biederman1987recognition}, hyperquadric~\cite{han1993using}, or superquadric~\cite{terzopoulos1991dynamic}.
To model the bi-ventricular shape, we pick a specific kind of superquadric, an ellipsoid, as our primitive:
\begin{equation}
    \boldsymbol{e}_{e}=a_{0}\begin{pmatrix}
a_{1}\, cos\, u\, cos\, v\\ 
a_{2}\, cos\, u\, sin\, v\\ 
a_{3}\, sin\, u\\ 
\end{pmatrix},
\end{equation}
where $-\pi /2\leq u\leq \alpha_{w}$, $-\pi \leq v <  \pi$, $a_{0}> 0$, $0\leq a_{1},a_{2},a_{3}< 1$. Specifically, $a_{0}$ is a scale parameter, and $a_{1}$, $a_{2}$ and $a_{3}$ are the aspect ratio parameters along the $x$-, $y$- and $z$-axes, respectively. We show an example ellipsoid in Fig.~\ref{fig3} (a).
To create a deformable shape primitive with more intuitive deformation degrees of freedom, we replace the constant parameters in a superquadric ellipsoid with \textit{parameter functions}~\cite{park1996deformable} of $u, w$:
\begin{equation}
\begin{aligned}
    \boldsymbol{e}&=\boldsymbol{e}(\boldsymbol{m};a_{0}(w), a_{1}(u, w), a_{2}(u, w), a_{3}(u, w) )
\\&=
a_{0}(w)\begin{pmatrix}
a_{1}(u, w)\, cos\, u\, cos\, v\\ 
a_{2}(u, w)\, cos\, u\, sin\, v\\ 
a_{3}(u, w)\, sin\, u\\ 
\end{pmatrix}.
\end{aligned}
\label{eq_pf}
\end{equation}

\subsection{Shape Blending for Bi-Ventricular Geometry}
As shown in Fig.~\ref{fig3} (b), the LV endo- and epi-cardial surfaces $\boldsymbol{s}(u, v, w=0,1)$ could be modeled by a two-layer deformable shape, defined in Eq.~(\ref{eq_pf}), with $w=0, 1$. Considering the significant shape difference of RV from LV, we use \textit{a blended shape} to model the RV. A single blended shape is the combination of two component shape primitive parts~\cite{decarlo1998shape}. We cut out portions of component primitives and join the remaining portions together:
\begin{equation}
    \boldsymbol{s}(u, v, 2)=\left\{\begin{matrix}
\boldsymbol{e}((u, v, 2);.,.,a_{2}(u, 2),.)\:\:   if\:\:   0\leq v< \pi \\ 
\boldsymbol{e}((u, -v, 3);.,.,a_{2}(u, 3),.)\:\:   if\:\:   -\pi\leq v< 0
\end{matrix}\right.,
\end{equation}
where $\boldsymbol{e}(u, v, 2)$ and $\boldsymbol{e}(u, v, 3)$ only differ by $a_{2}(u)$. In this way, we model the bi-ventricular geometry as blended surfaces $\boldsymbol{s}(u, v, w)$ with $w=0,1,2$,  $\alpha_{0}=\alpha_{1}=\pi/6$, $\alpha_{2}=\pi$. 

Although (geometry) parameter functions greatly increase the capturing capacity of shape features beyond constant parameter-based superquadrics, we can further apply global deformations to the underlying shape primitive $\boldsymbol{e}$ with continuous  deformation parameter functions. Here, we apply axis offset deformations $\mathbf{T}_{o}$ to $\boldsymbol{e}$:
\begin{equation}
    \boldsymbol{s}=\mathbf{T}_{o}(\boldsymbol{e}; e_{xo}(u), e_{yo}(u))=\begin{pmatrix}
e_{x}+e_{xo}(u, w)\\ 
e_{y}+e_{yo}(u, w)\\ 
e_{z}
\end{pmatrix},
\end{equation}
where $e_{xo}(u, w)$, $e_{yo}(u, w)$ are axis-offset parameter functions along the $x$- and $y$- axes, respectively. We show an example of $\mathbf{T}_{o}$ deformed ellipsoid in Fig.~\ref{fig3} (c). It is obvious that $\mathbf{T}_{o}$ describes the bending deformation along the long axis of the heart ventricle. Note that, although the parameter functions are defined with local $u$, they are of ``global'' deformation effect on each $u$. Therefore, we call them global parameters in our work.

\subsection{Diffeomorphic Point Flow for Local Deformations}
In Eq.~(\ref{eq_qcsd}), we use local deformation $\boldsymbol{d}$ to describe details of complex shapes. In our NDM, local deformation is modeled as diffeomorphic point flows. Let $\mathcal{D}(\boldsymbol{q}, t):\mathcal{V} \subset \mathbb{R}^{N\times 3}\times [0, 1] \mapsto \mathcal{V} \subset \mathbb{R}^{N\times 3}$ be a time $t\in \left [ 0,1 \right ]$ parameterized spatial mapping such that, for $N$ points $\boldsymbol{q}=\boldsymbol{q}^{(0)}=\mathcal{D}(\boldsymbol{q}, 0)$, ${\boldsymbol{q}}'=\boldsymbol{q}^{(1)}=\mathcal{D}(\boldsymbol{q}, 1)$ are the final transformed points, and $\boldsymbol{d}=\mathcal{D}(\boldsymbol{q}, 1)-\mathcal{D}(\boldsymbol{q}, 0)$. A diffeomorphic mapping is achieved as the trajectory and integration of a smooth \textit{velocity field} $\boldsymbol{v}:\mathcal{V} \subset \mathbb{R}^{N\times 3}\times [0, 1] \mapsto \mathcal{V} \subset \mathbb{R}^{N\times 3}$, which is governed by an ODE, also known as the flow equation~\cite{dupuis1998variational}, as follows: 
\begin{equation}
    \frac{\partial \mathcal{D}(\boldsymbol{q}, t)}{\partial t}=\boldsymbol{v}(\mathcal{D}(\boldsymbol{q}, t), t)\; \;\; \; s.t.\; \;\; \; \mathcal{D}(\boldsymbol{q}, 0)=\boldsymbol{q}.
\label{eq_flow}
\end{equation}
The initial value problem (IVP) in Eq.~(\ref{eq_flow}) could be solved with a neural ODE solver~\cite{chen2018neural} with the dynamic function the velocity field. According to the \textit{Cauchy-Lipschitz} theorem~\cite{brezis2011functional}, if the velocity field is Lipschitz continuous, the resulting transform $\mathcal{D}$ is a \textit{bi-Lipschitz map}, which is also a diffeomorphism in essence~\cite{dupont2019augmented}.

In this work, we consider the pose ($\boldsymbol{c}$ and $\boldsymbol{R}$) and shape parameter functions in the primitive $\boldsymbol{s}$ as generalized deformations. Therefore, our neural deformable model consists of the following deformation parameter vector:
\begin{equation}
    \boldsymbol{q}_{N}=(\boldsymbol{q}_{g}^{\top}, \boldsymbol{q}_{d}^{\top})^{\top},
\end{equation}
where $\boldsymbol{q}_{g}= (\boldsymbol{c}^{\top}, \boldsymbol{R}^{\top}, \boldsymbol{a}_{0}, \boldsymbol{a}_{1}, \boldsymbol{a}_{2}, \boldsymbol{a}_{3}, \boldsymbol{e}_{xo}, \boldsymbol{e}_{yo})^{\top}$ is the global deformation parameter vector and $\boldsymbol{q}_{d}= \boldsymbol{d}$ is the local deformation parameter vector.

\begin{figure*}
\begin{center}
\includegraphics[width=1.0\linewidth]{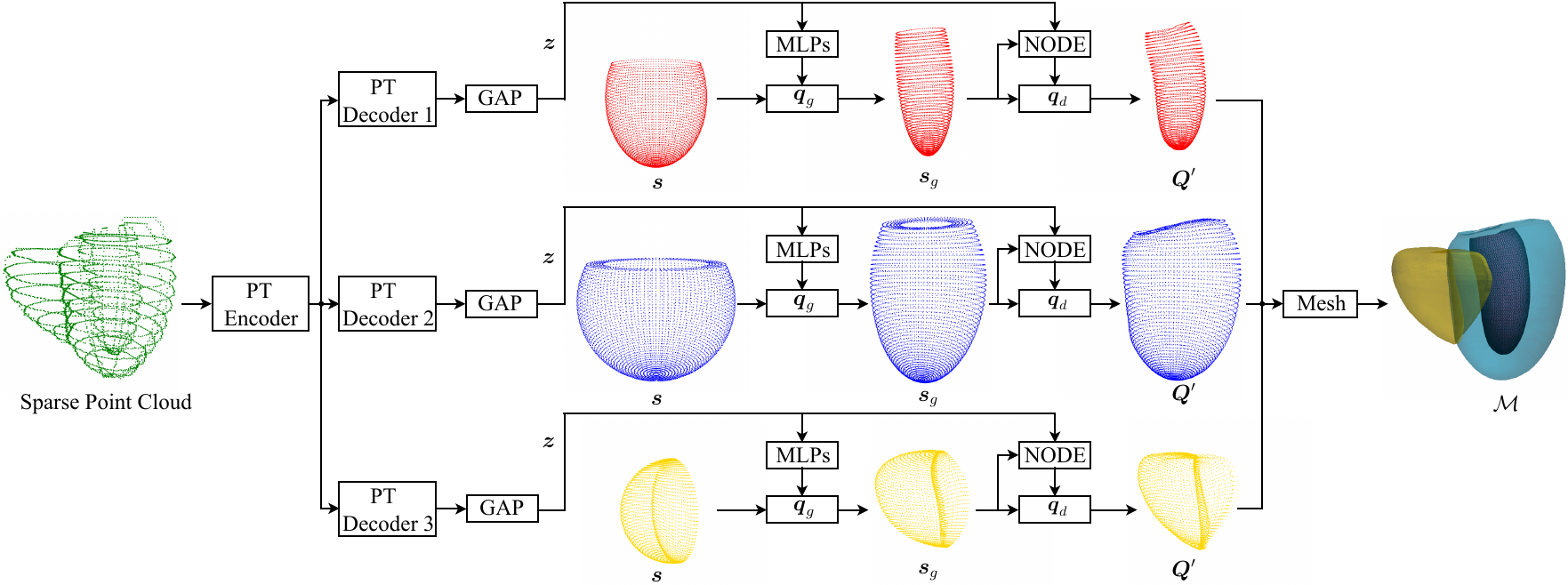}
\end{center}
   \caption{Neural deformable models (NDMs) for bi-ventricular cardiac point cloud upsampling and automated triangular mesh generation. Each deformation parameter vector $\boldsymbol{q}_{N}=(\boldsymbol{q}_{g}^{\top}, \boldsymbol{q}_{d}^{\top})^{\top}$ of a NDM is learned in a coarse-to-fine fashion, from the learning of global deformation ($\boldsymbol{q}_{g}$) to local deformation ($\boldsymbol{q}_{d}$). $\boldsymbol{q}_{g}$ captures gross target shape features, leaving only small shape details to be refined by $\boldsymbol{q}_{d}$.}
\label{fig4}
\end{figure*}

\subsection{Architecture of NDM}
We learn $\boldsymbol{q}_{N}$ for each shape instance in a coarse-to-fine fashion: we first learn $\boldsymbol{q}_{g}$ and then we learn $\boldsymbol{q}_{d}$. As shown in Fig.~\ref{fig4}, our NDM has three branches that predict each $\boldsymbol{q}_{N}$ of the LV endo-, epi-cardial surfaces and the RV surface, respectively. We use a shared point transformer (PT) encoder and three point transformer decoders to learn shape embeddings from a given sparse point cloud. The PT architecture is the same as in~\cite{zhao2021point} for semantic segmentation, except that we replace its last multi-layer perceptron (MLP) layer with a global average pooling (GAP) layer to get global shape embedding $\boldsymbol{z}$.
With $\boldsymbol{z}$, we first use MLPs to predict the global deformation parameter vector $\boldsymbol{q}_{g}$. The shape primitive $\boldsymbol{s}$ is then deformed by $\boldsymbol{q}_{g}$:
\begin{equation}
    \boldsymbol{s}_{g}=\boldsymbol{s}\circ\boldsymbol{q}_{g},
\end{equation}
where $\circ$ describes global deformations, and $\boldsymbol{s}_{g}$ is the resulting globally deformed shape primitive, which captures the coarse target shape features, as shown in Fig.~\ref{fig4}. As we will illustrate next, during training of NDM, we exploit a marginal space learning method~\cite{zheng2007fast} to enforce the global deformations to account for as much of the target shape as possible.
Then we use a \textit{conditional} diffeomorphic point flow generation block $\mathcal{D}$ to learn local deformations:
\begin{equation}
    \frac{\partial \mathcal{D}(\boldsymbol{q};\boldsymbol{z}, t)}{\partial t}=\boldsymbol{v}(\mathcal{D}(\boldsymbol{q}, t);\boldsymbol{z}, t)\; \;\; \; s.t.\; \;\; \; \mathcal{D}(\boldsymbol{q};\boldsymbol{z}, 0)=\boldsymbol{s}_{g}.
\label{cfloweq}
\end{equation}
We call $\mathcal{D}$ the neural ordinary equation (NODE) block. We use the same NODE block as in~\cite{gupta2020neural}, except that we exclude its instance normalization layer. By solving Eq.~(\ref{cfloweq}), we get the final reconstructed shape ${\boldsymbol{Q}}'=\mathcal{D}(\boldsymbol{q};\boldsymbol{z}, 1)$. From Fig.~\ref{fig4}, the local deformations can further refine target shape details. As such, our coarse-to-fine architecture of NDM results in accurate target shape reconstructions.

\subsection{Triangular Mesh Generation}
Both global and local deformations in $\boldsymbol{q}_{N}$ are smooth with inverse smooth. Therefore, $\boldsymbol{q}_{N}$ is a diffeomorphic mapping and the final shape reconstruction result ${\boldsymbol{Q}}'$ preserves the topology of shape primitive $\boldsymbol{s}$. Since $\boldsymbol{s}$ is an ellipsoid, we can get the corresponding mesh by connecting any three nearest-neighboring surface vertices. We then take the edges of this ellipsoid mesh as those of our target mesh. In this way, we can automatically reconstruct a triangular mesh $\mathcal{M}$ from ${\boldsymbol{Q}}'$. 

\begin{figure}[t]
\begin{center}
\includegraphics[width=0.5\linewidth]{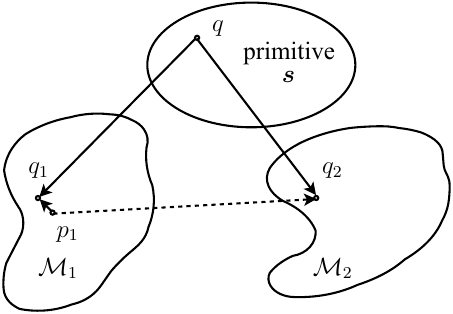}
\end{center}
   \caption{Shape registration via the implicitly learned dense correspondence by NDM. }
\label{fig5}
\end{figure}

\subsection{Shape Registration}
NDM implicitly learns dense correspondences between different shape instances, which can be used to register shapes. As shown in Fig.~\ref{fig5}, we first use NDM to fit the primitive $\boldsymbol{s}$ to the targets $\mathcal{M}_{1}$ and $\mathcal{M}_{2}$. The dense correspondences learned by NDM will map a point $q$ of the primitive to $q_{1}$ and $q_{2}$ for $\mathcal{M}_{1}$ and $\mathcal{M}_{2}$, respectively. Obviously, $q_{1}$ may not coincide with the point $p_{1}$ of $\mathcal{M}_{1}$. For each point $p_{1}$ of $\mathcal{M}_{1}$, we first look up the nearest neighbouring point $q_{1}$ reconstructed by NDM, then we map $p_{1}$ to $q_{2}$ via the correspondence $p_{1}\rightarrow q_{1}\rightarrow q\rightarrow q_{2}$. In this way, we can register $\mathcal{M}_{1}$ to $\mathcal{M}_{2}$.

\subsection{Model Training}
We use the Chamfer distance (CD) loss~\cite{fan2017point} to encourage the geometric similarity between vertices ${\boldsymbol{Q}}'$ of the reconstructed mesh and the ground truth point cloud $\mathcal{Q}$:
\begin{equation}
    \mathcal{L}_{geo}=\mathcal{L}_{CD}(\boldsymbol{Q}', \mathcal{Q}).
\end{equation}
We also include two regularization losses. The first one is to regularize the amount of local deformations:
% \begin{equation}
%     \mathcal{L}_{d}=\frac{1}{N}\sum_{i=1}^{N}\left \| d_{i} \right \|^{2},
% \end{equation}
\begin{equation}
    \mathcal{L}_{d}=\left \| \boldsymbol{q}_{d} \right \|_{2}^{2},
\end{equation}
where  $\boldsymbol{q}_{d}=\boldsymbol{Q}' - \boldsymbol{s}_{g}$ is the local deformation field. The second one is to regularize the smoothness of the local deformation field:
\begin{equation}
    \mathcal{L}_{s}=\left \| \bigtriangledown  \boldsymbol{q}_{d} \right \|_{2}^{2},
\end{equation}
where $\bigtriangledown \boldsymbol{q}_{d}$ is the gradients of $\boldsymbol{q}_{d}$ over the material coordinates $u$ and $v$. In summary, the total loss function is the weighted sum of the geometric similarity loss and the local deformation regularization terms:
\begin{equation}
    \mathcal{L}=\mathcal{L}_{geo}+\lambda_{d} \mathcal{L}_{d}+\lambda_{s} \mathcal{L}_{s},
\label{loss_gds}
\end{equation}
where $\lambda_{d}$ and $\lambda_{s}$ are the weighting hyper-parameters.

The learning of $\boldsymbol{q}_{N}$ is to search the optimum deformation parameters. The search space composed of $\boldsymbol{q}_{N}$ is so large that end-to-end training by optimizing Eq.~(\ref{loss_gds}) leads to slow convergence. We adopt the marginal space learning (MSL) method~\cite{zheng2007fast, raju2022deep} to train NDM. More specifically, we decompose the training process into a chain of sub-processes, in which we gradually add one component of the deformation parameter vector $\boldsymbol{q}_{N}$ into NDM at a time.
In this way, each training sub-process only focuses on one individual deformation component, to speed up the convergence.

\section{Experiments}
\subsection{Dataset and Pre-Processing}
We use a large public 3D CMR dataset~\cite{bai2015bi} of $1,331$ normal subjects to evaluate our method. Each subject contains the end-diastole (ED) and end-systole (ES) phases.
Both the low resolution (LR) and high resolution (HR) segmentation masks of each subject are provided in this dataset. 
We took the HR data, which was scanned by 3D high spatial resolution of $1.25\times1.25\times2\; mm^{3}$ CMR protocols~\cite{de2014population} as our experimental dataset.
We extracted vertices from the bi-ventricular mesh reconstructed from the segmentation mask volume to form a dense ground truth point cloud. 
Next, we generated a sparse point cloud as the input by mimicking standard clinical 2D CMR scanning. We first determined the LAX and SAX view planes. 
Then we sliced the segmentation mask volume at these planes and extracted points along the segmentation contours. We downsampled these points using farthest point sampling (FPS)~\cite{qi2017pointnet++} to a fixed number. 
We sampled 3 LAX and 10 SAX planes and we set the input point number as $5,600$. 
We randomly split the dataset into 900, 200 and 231 subjects as the training, validation and test sets, respectively. 
For the pre-processing, we centered each input point cloud at $(0, 0, 0)$ by subtracting the center coordinates and linearly normalized $x$, $y$, $z$ coordinates to $[-0.85, 0.85]$. We also rotated each data set to align the center line of LV and RV with the $y$-axis.

\subsection{Evaluation Metrics}
For quantitative evaluation of the geometric similarity, we followed~\cite{chen2021shape, li2021point} to use Chamfer distance (CD), earth mover’s distance (EMD) and point-to-surface distance (P2F) to measure the geometric similarity between the densified point cloud and the ground truth. 
For the evaluation of generated mesh quality, we followed~\cite{sun2022topology} to compute the normal consistency (NC), easy non-manifold face (ENF) ratio (with $\delta=0$), and self-intersection (SI) ratio.
We also used CD, EMD, P2F to measure the shape registration accuracy.
The lower the evaluation metric, the better the performance, except for NC, which is the opposite. 
Note that we do all the quantitative evaluation in the original data space, not in the normalized data space.

\subsection{Baseline Methods}
We compared our method with a state-of-the-art (SOTA) 3D bi-ventricular geometry reconstruction method MR-Net~\cite{chen2021shape}, and a SOTA 3D manifold mesh generation method NMF~\cite{gupta2020neural}. 
For MR-Net, we first randomly sampled a data set from the training dataset; then we used our method to fit the primitive to the ground truth shape; the reconstructed shape was used as the template mesh. 
For NMF, the template spheres were replaced by the same initial shape primitives as ours.
We used their publicly available codes and followed the same setting of the original work to train their networks with our training dataset. 
Note that we computed the geometric similarity losses of LV-endo, LV-epi and RV surfaces separately for the training of our method, MR-Net and NMF.
We also trained our model in an unsupervised fashion (Ours-un), with the input sparse point cloud as the ground truth, which is similar to conventional iterative optimization-based deformable models.
We compared our accuracy of shape registration based on the learned dense correspondence with MR-Net, NMF and Ours-un.

\subsection{Implementation Details}
We implemented NDM with PyTorch. And we set the hyper-parameters $\lambda_{d}=0.1$, $\lambda_{s}=0.05$. Adam optimizer is used with a learning rate of $5e^{-4}$ to train our model with a batch size of 2. During training, we sampled $5,000$, $5,500$ and $5,000$ points from the reconstructed LV-endo, LV-epi and RV meshes, respectively. All models were trained on an NVIDIA Quadro RTX 8000 GPU. For evaluation, we uniformly sampled the same number of $3,000$ points, on the predicted mesh as the ground truth for the calculation of CD, EMD and P2F for the LV-endo, LV-epi and RV surfaces, respectively.

\subsection{Results}

\begin{table}[t]
\begin{center}
\resizebox{1.0\columnwidth}{!}{
\begin{tabular}{c|l|c|c|c|c|c|c}
\hline
Phase &Method &CD $\downarrow$ &EMD $\downarrow$ &P2F $\downarrow$ &NC $\uparrow$&ENF $\downarrow$ &SI $\downarrow$\\
\hline\hline
\multirow{5}{*}{ED} &MR-Net &$3.09$&$6.05$&$1.47$&$0.762$&$1.48$&$0.035$\\
 &NMF &$6.31$&$6.50$&$4.40$&$0.759$&$1.46$&$0.019$\\
 &Ours-un &$5.16$&$6.20$&$3.11$&$0.751$&$\mathbf{1.45}$&$0.002$\\
&Ours &$\mathbf{2.73}$&$\mathbf{5.80}$&$\mathbf{1.17}$&$\mathbf{0.765}$&$1.46$&$\mathbf{0}$\\
 \hline\hline
 \multirow{5}{*}{ES} &MR-Net &$2.37$&$4.50$&$1.23$&$0.750$&$1.47$&$0.068$\\
 &NMF &$3.74$&$4.74$&$2.46$&$0.752$&$1.46$&$0.027$\\
 &Ours-un &$2.90$&$4.48$&$1.63$&$0.747$&$\mathbf{1.45}$&$0.002$\\
 &Ours &$\mathbf{1.91}$&$\mathbf{4.07}$&$\mathbf{0.873}$&$\mathbf{0.758}$&$1.46$&$\mathbf{0}$\\

\hline
\end{tabular}}
\end{center}
\caption{Mean Chamfer distance (CD) ($mm$), earth mover’s distance (EMD) ($mm$) and point-to-surface distance (P2F) ($mm$) for geometric similarity evaluation. Mean normal consistency (NC), easy non-manifold face (ENF) ratio and self-intersection (SI) ratio for cardiac mesh quality evaluation.}
\label{table1}
\end{table}

\subsubsection{Shape Reconstruction Performance}
In Table~\ref{table1}, we show the mean quantitative results of bi-ventricle shape reconstruction for ED and ES phases. More detailed quantitative results of LV-endo, LV-epi and RV surface reconstruction are presented in Supplementary Material. We also show an example in Fig.~\ref{fig6}.
Our method significantly outperforms baseline methods for both geometric similarity and mesh quality aspects. MR-Net cannot handle large shape variations between the template mesh and the target, resulting in significant shape artefacts in the reconstructions. Such artefacts could produce self-intersected local surfaces, as demonstrated by the SI value. NMF utilizes multiple deformation blocks to map the template to the target. However, such multiple deformation method can only learn coarse shape features. It cannot recover complex shape details, which gives the worst geometric similarity between the reconstruction and the ground truth. Ours-un method only makes use of observed visual data, which is at the risk of overfitting to sparse observation. Our NDM method deals with shape reconstruction in a coarse-to-fine fashion and can learn the implicit shape correspondence from a shape manifold. Therefore, it can accurately reconstruct the bi-ventricular shape and generate high quality meshes.

\begin{table}[t]
\begin{center}
\resizebox{0.9\columnwidth}{!}{
\begin{tabular}{l|c|c|c}
\hline
Method &CD ($mm$) $\downarrow$ &EMD ($mm$) $\downarrow$ &P2F ($mm$) $\downarrow$ \\
\hline\hline
MR-Net &$2.44$&$4.74$&$1.33$\\
NMF &$3.76$&$4.92$&$2.45$\\
Ours-un &$3.03$&$4.41$&$1.91$\\
Ours &$\mathbf{1.94}$&$\mathbf{3.95}$&$\mathbf{1.00}$\\
\hline
\end{tabular}}
\end{center}
\caption{Mean Chamfer distance (CD), earth mover’s distance (EMD) and point-to-surface distance (P2F) for shape registration.}
\label{table2}
\end{table}

\begin{figure*}
\begin{center}
\includegraphics[width=1.0\linewidth]{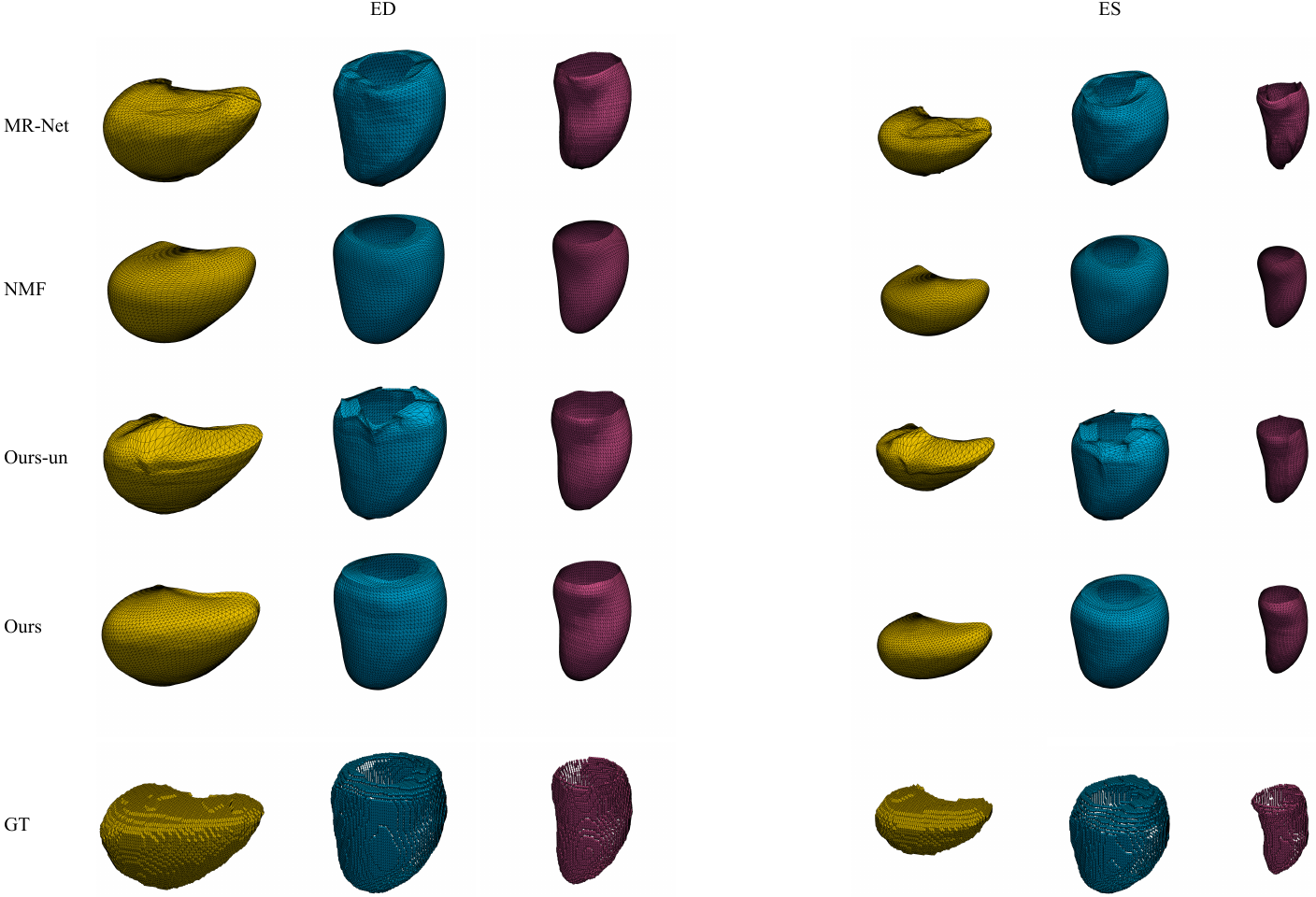}
\end{center}
   \caption{Bi-ventricle shape reconstruction results. For each method, we show results of both the ED and ES phases, in which the LV endo-cardial, epi-cardial and RV surfaces are shown in red, blue and yellow, respectively.}
\label{fig6}
\end{figure*}

\begin{figure}
\begin{center}
\includegraphics[width=1.0\linewidth]{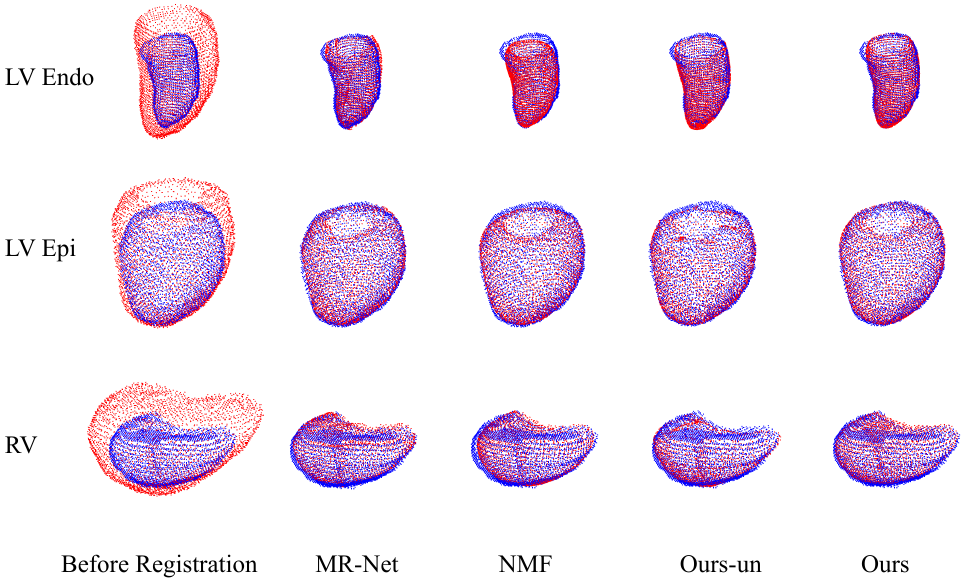}
\end{center}
   \caption{Bi-ventricle shape registration results. For each method, we show registration result from the ED phase to the ES phase, in which the (warped) ED and ES phases are shown in red and blue, respectively.}
\label{fig7}
\end{figure}

\begin{figure*}
\begin{center}
\includegraphics[width=1.0\linewidth]{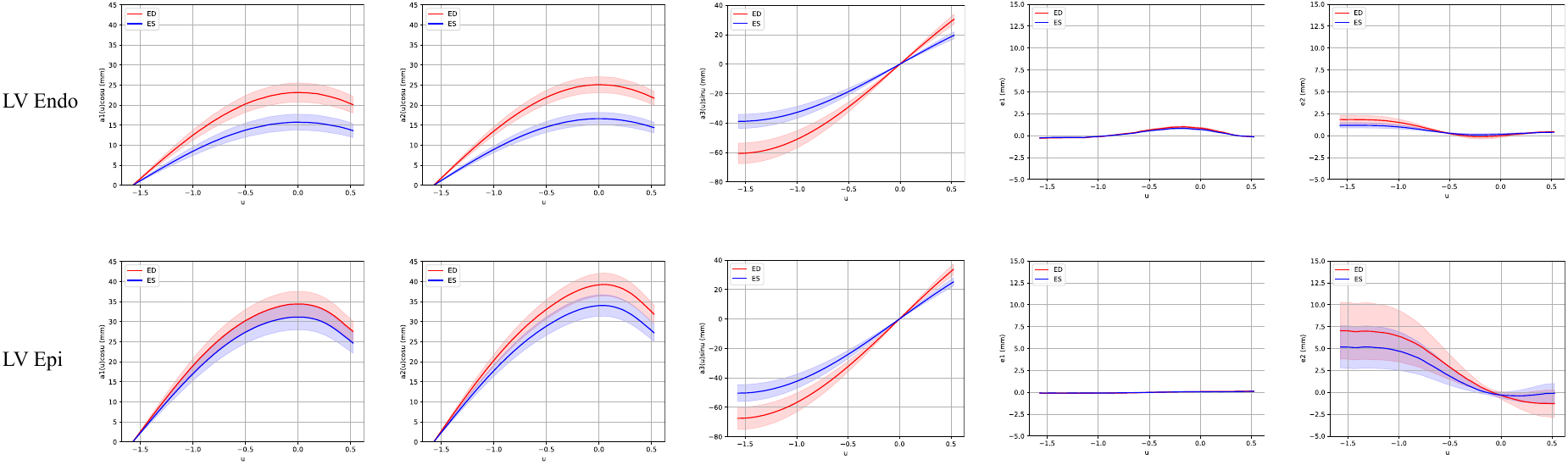}
\end{center}
   \caption{Learned left ventricle geometric and global long axis off set parameters by NDM. Curves show the mean values; color bands show the standard deviations.}
\label{fig8}
\end{figure*}

\subsubsection{Shape Registration Performance}
In Table~\ref{table2}, we show the mean quantitative results of shape registration from ED phase to ES phase. More detailed quantitative results of LV-endo, LV-epi and RV surface registration are presented in Supplementary Material. We also show an example in Fig.~\ref{fig7}. The shape registration is achieved via the learned dense shape correspondence. Thus, the shape registration accuracy reflects the accuracy of implicitly learned dense correspondence. Our method outperforms all baseline methods by a large margin. The results indicate that our method has the potential of achieving accurate heart motion tracking. Note that our method learns 3D dense correspondence between heart shape instances from information extracted from 2D image stacks. This is of great importance for the 3D heart motion field estimation for 2D-based cardiac MR imaging.

\subsubsection{Intuitive Interpretation of NDM Parameters}
NDM is not only powerful for shape reconstruction and registration, but also enables intuitive interpretation of the heart geometry and deformation parameters, which is a distinct property not shared by any baseline methods. In 
 Fig.~\ref{fig8}, we show the five different kinds of parameters of the left ventricle (LV) endo- and epi-cardial surfaces calculated on the test dataset. We can obtain intuitive information from these plots. The aspect ratios ${a}'_{1}(u)=a_{1}(u)\, cos\, u$ and ${a}'_{2}(u)=a_{2}(u)\, cos\, u$ describe the short axis LV ``radius'' distribution range along the long axis of the heart. The aspect ratio ${a}'_{3}(u)=a_{3}(u)\, sin\, u$ describes the long axis LV ``radius'' distribution range along the long axis of the heart. The axis offsets $e_{1}(u)$ ($e_{xo}(u)$) and $e_{2}(u)$ ($e_{yo}(u)$) describe the long axis bending information. Importantly, all of these parameters can be directly used by clinicians without any complex post-processing. We can also derive other useful information about the LV from these basic parameters, based on multiple temporal data points. For example, by comparing ${a}'_{1}$,  ${a}'_{2}$,  ${a}'_{3}$ for ED phase and ES phase, we get an LV contraction metric. By studying the metrics based on populations, we can immediately build an LV myocardium wall motion atlas.

\begin{figure}
\begin{center}
\includegraphics[width=1.0\linewidth]{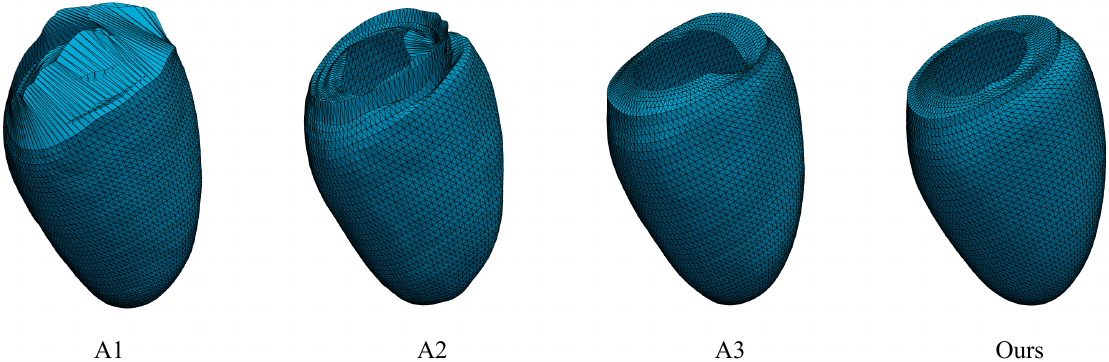}
\end{center}
   \caption{Ablation study results of the LV-epi surface reconstruction. Note the shape reconstruction differences from different models near the basal region. }
\label{fig9}
\end{figure}

\begin{table}[t]
\begin{center}
\resizebox{0.7\columnwidth}{!}{
\begin{tabular}{l|c|c|c|c}
\hline
Model &$\mathcal{L}_{d}$ &$\mathcal{L}_{s}$ &CD ($mm$) $\downarrow$&SI $\downarrow$\\
\hline\hline
A1 &\xmark & \xmark &$4.49$ &$0.049$\\
A2 &\cmark & \xmark &$3.15$ &$0.049$\\
A3 &\xmark & \cmark  &$2.88$ &$0.001$\\
Ours &\cmark & \cmark &$\mathbf{2.73}$ &$\mathbf{0}$\\

\hline
\end{tabular}}
\end{center}
\caption{Ablation of the local deformation regularization terms in NDM on ED phase data.}
\label{table3}
\end{table}

\subsubsection{Ablation Study}
In Table~\ref{table3}, we present the effects of local deformation regularization terms proposed in our NDM model. We also show an example of LV-epi surface reconstruction in Fig.~\ref{fig9}. Our method is different from MR-Net, which uses graph convolutional networks and mesh-based metrics for the deformation regularization. We use NODE to learn a diffeomorphism point flow. The advantage of this method is that we can get rid of complex geometric computation in order to learn plausible shape reconstructions. However, as shown by model A1, A2 and A3, we still need explicit regularization to achieve the goal. The reconstruction differences of these models are manifested mainly around the basal region, where distinct shape variations are observed. Without the explicit local deformation regularization terms, even the use of NODE could lead to unpleasant shape reconstruction overfitting. However, based on our novel design of the NDM model, we can use simple local deformation total amount ($\mathcal{L}_{d}$) and neighbouring smoothness ($\mathcal{L}_{s}$) regularization to achieve accurate shape reconstructions.

\section{Conclusion}
In this work, we proposed a novel neural deformable model to reconstruct bi-ventricular shape from sparse visual data. We inherited a conventional parameter function-based deformable model and incorporated neural ordinary differential equation blocks for local diffeomorphic deformations. We designed a coarse-to-fine learning paradigm to make NDM successfully learn from a shape distribution manifold. Based on such a learning paradigm, only using a simple local deformation regularization could overcome potential overfitting problems. Our NDM model can not only reconstruct bi-ventricular shape accurately, but also enable accurate 3D shape registration. More importantly, our NDM models provide clinicians with intuitive heart shape and deformation parameters in a straightforward manner.

{\small
\bibliographystyle{ieee_fullname}
\bibliography{main}
}

%\clearpage
\section{Supplementary Material}
\subsection{Detailed Quantitative Results of Bi-ventricular Shape Reconstruction}
In Table~\ref{tables1}, we show the detailed Chamfer distance (CD), earth mover’s distance (EMD) and point-to-surface distance (P2F) for geometric similarity evaluation of LV-endo, LV-epi and RV shape reconstructions. Our method outperforms all the baseline methods for LV-endo, LV-epi and RV shape reconstruction, respectively.

In Table~\ref{tables2}, we show the detailed normal consistency (NC), easy non-manifold face (ENF) ratio and self-intersection (SI) ratio for reconstructed mesh quality evaluation of LV-endo, LV-epi and RV shapes.
On average, our method can generate better cardiac meshes than all the baseline methods.

\begin{table*}[t]
\begin{center}
\resizebox{2.0\columnwidth}{!}{
\begin{tabular}{l|c|c|c|c|c|c|c|c|c|c|c|c|c}
\hline
\multirow{2}{*}{Phase} &\multirow{2}{*}{Method} &\multicolumn{4}{c}{CD ($mm$) $\downarrow$} &\multicolumn{4}{|c|}{EMD ($mm$) $\downarrow$} &\multicolumn{4}{|c}{P2F ($mm$) $\downarrow$}\\
\cline{3-14}
 & & Endo & Epi& RV & avg & Endo & Epi& RV & avg & Endo & Epi& RV & avg\\
\hline\hline
\multirow{5}{*}{ED} &MR-Net &$2.56\pm1.14$ &$3.32\pm1.10$&$3.39\pm1.07$&$3.09\pm1.06$&$6.04\pm1.19$&$6.67\pm1.71$&$5.44\pm1.67$&$6.05\pm1.17$&$1.26\pm0.984$&$1.32\pm0.817$&$1.82\pm0.945$&$1.47\pm0.877$\\
 &NMF &$5.85\pm4.43$ &$5.67\pm4.33$&$7.40\pm8.96$&$6.31\pm5.54$&$6.16\pm1.37$&$6.61\pm1.68$&$6.73\pm2.27$&$6.50\pm1.43$&$4.07\pm3.70$&$3.45\pm3.88$&$5.69\pm8.03$&$4.40\pm4.90$\\
 &Ours-un &$2.72\pm0.915$ &$4.38\pm0.978$&$8.38\pm4.46$&$5.16\pm1.70$&$5.93\pm1.31$&$6.42\pm1.60$&$6.24\pm1.99$&$6.20\pm1.26$&$1.37\pm0.704$&$2.15\pm0.682$&$5.81\pm3.56$&$3.11\pm1.30$\\
&Ours &$\mathbf{2.25}\pm1.03$ &$\mathbf{2.91}\pm0.762$&$\mathbf{3.02}\pm0.892$&$\mathbf{2.73}\pm0.795$&$\mathbf{5.70}\pm1.14$&$\mathbf{6.38}\pm1.72$&$\mathbf{5.33}\pm1.69$&$\mathbf{5.80}\pm1.17$&$\mathbf{1.01}\pm0.850$&$\mathbf{1.06}\pm0.515$&$\mathbf{1.43}\pm0.712$&$\mathbf{1.17}\pm0.606$\\
 \hline\hline
 \multirow{5}{*}{ES} &MR-Net &$2.11\pm0.898$ &$2.73\pm0.755$&$2.28\pm0.805$&$2.37\pm0.729$&$2.59\pm0.524$&$5.27\pm1.53$&$5.64\pm1.63$&$4.50\pm0.666$&$1.18\pm0.801$&$1.16\pm0.539$&$1.35\pm0.767$&$1.23\pm0.641$\\
 &NMF &$3.41\pm3.10$ &$3.74\pm1.47$&$4.05\pm2.83$&$3.74\pm2.26$&$2.76\pm0.870$&$5.29\pm1.38$&$6.17\pm1.72$&$4.74\pm0.796$&$2.25\pm2.35$&$2.04\pm1.13$&$3.11\pm2.59$&$2.46\pm1.86$\\
 &Ours-un &$1.97\pm0.631$ &$3.44\pm0.663$&$3.27\pm1.55$&$2.90\pm0.732$&$2.93\pm0.633$&$5.14\pm1.27$&$5.37\pm1.56$&$4.48\pm0.656$&$1.15\pm0.483$&$1.77\pm0.433$&$1.95\pm1.13$&$1.63\pm0.516$\\
 &Ours &$\mathbf{1.50}\pm0.542$ &$\mathbf{2.32}\pm0.582$&$\mathbf{1.92}\pm0.823$&$\mathbf{1.91}\pm0.556$&$\mathbf{2.23}\pm0.481$&$\mathbf{4.92}\pm1.42$&$\mathbf{5.06}\pm1.49$&$\mathbf{4.07}\pm0.607$&$\mathbf{0.759}\pm0.397$&$\mathbf{0.914}\pm0.356$&$\mathbf{0.947}\pm0.605$&$\mathbf{0.873}\pm0.392$\\

\hline
\end{tabular}}
\end{center}
\caption{Detailed Chamfer distance (CD), earth mover’s distance (EMD) and point-to-surface distance (P2F) for geometric similarity evaluation of LV-endo, LV-epi and RV shape reconstruction, respectively.}
\label{tables1}
\end{table*}

%%%%%%%%%%%%%%%%%%%%%%%%%%%%%%%%%%%%%%%%%%%%%%%%%
%%%%%%%%%%%%%%%%%%%%%%%%%%%%%%%%%%%%%%%%%%%%%%%%%
%%%%%%%%%%%%%%%%%%%%%%%%%%%%%%%%%%%%%%%%%%%%%%%%%

\begin{table*}[t]
\begin{center}
\resizebox{2.0\columnwidth}{!}{
\begin{tabular}{l|c|c|c|c|c|c|c|c|c|c|c|c|c}
\hline
\multirow{2}{*}{Phase} &\multirow{2}{*}{Method} &\multicolumn{4}{c}{NC $\uparrow$} &\multicolumn{4}{|c|}{ENF $\downarrow$} &\multicolumn{4}{|c}{SI $\downarrow$}\\
\cline{3-14}
 & & Endo & Epi& RV & avg & Endo & Epi& RV & avg & Endo & Epi& RV & avg\\
\hline\hline
\multirow{5}{*}{ED} &MR-Net &$0.713\pm0.034$ &$0.707\pm0.033$&$0.866\pm0.012$&$0.762\pm0.024$&$1.48\pm0.004$&$1.48\pm0.002$&$1.47\pm0.004$&$1.48\pm0.003$&$0.035\pm0.024$&$0.016\pm0.012$&$0.055\pm0.018$&$0.035\pm0.013$\\
 &NMF &$0.715\pm0.031$ &$0.707\pm0.033$&$0.854\pm0.024$&$0.759\pm0.024$&$1.47\pm0$&$1.47\pm0$&$1.44\pm0$&$1.46\pm0$&$0.011\pm0.027$&$0.016\pm0.026$&$0.030\pm0.046$&$0.019\pm0.020$\\
 &Ours-un &$\mathbf{0.717}\pm0.033$ &$0.701\pm0.032$&$0.835\pm0.016$&$0.751\pm0.022$&$1.47\pm0$&$1.47\pm0.002$&$\textbf{1.42}\pm0.004$&$\mathbf{1.45}\pm0.002$&$0\pm0$&$0\pm0.001$&$0.007\pm0.011$&$0.002\pm0.004$\\
&Ours &$0.716\pm0.032$ &$\mathbf{0.709}\pm0.034$&$\textbf{0.871}\pm0.011$&$\mathbf{0.765}\pm0.023$&$\mathbf{1.47}\pm0$&$\mathbf{1.47}\pm0$&$1.44\pm$0&$1.46\pm0$&$\mathbf{0}\pm0$&$\mathbf{0}\pm0.001$&$\mathbf{0}\pm0$&$\mathbf{0}\pm0$\\
 \hline\hline
 \multirow{5}{*}{ES} &MR-Net &$0.708\pm0.028$ &$0.698\pm0.033$&$0.843\pm0.017$&$0.750\pm0.022$&$\mathbf{1.46}\pm0.010$&$1.49\pm0.003$&$1.46\pm0.006$&$1.47\pm0.005$&$0.113\pm0.042$&$0.012\pm0.014$&$0.078\pm0.024$&$0.068\pm0.018$\\
 &NMF &$0.716\pm0.030$ &$0.699\pm0.034$&$0.840\pm0.021$&$0.752\pm0.023$&$1.47\pm0$&$1.47\pm0$&$1.44\pm0$&$1.46\pm0$&$0.012\pm0.025$&$0.018\pm0.025$&$0.051\pm0.082$&$0.027\pm0.030$\\
 &Ours-un &$\mathbf{0.719}\pm0.031$ &$0.693\pm0.033$&$0.830\pm0.015$&$0.747\pm0.022$&$1.47\pm0$&$1.47\pm0.001$&$\mathbf{1.43}\pm0.004$&$\mathbf{1.45}\pm0.001$&$0\pm0$&$0\pm0$&$0.004\pm0.006$&$0.002\pm0.002$\\
&Ours &$0.715\pm0.032$ &$\mathbf{0.701}\pm0.033$&$\mathbf{0.857}\pm0.013$&$\mathbf{0.758}\pm0.022$&$1.47\pm0$&$\mathbf{1.47}\pm0$&$1.44\pm0$&$1.46\pm0$&$\mathbf{0}\pm0$&$\mathbf{0}\pm0.004$&$\mathbf{0}\pm0.002$&$\mathbf{0}\pm0.001$\\

\hline
\end{tabular}}
\end{center}
\caption{Detailed normal consistency (NC), easy non-manifold face (ENF) ratio and self-intersection (SI) ratio for reconstructed mesh quality evaluation of LV-endo, LV-epi and RV shape, respectively.}
\label{tables2}
\end{table*}

%%%%%%%%%%%%%%%%%%%%%%%%%%%%%%%%%%%%%%%%%%%%%%%%%
%%%%%%%%%%%%%%%%%%%%%%%%%%%%%%%%%%%%%%%%%%%%%%%%%
%%%%%%%%%%%%%%%%%%%%%%%%%%%%%%%%%%%%%%%%%%%%%%%%%
\subsection{Detailed Quantitative Results of Bi-ventricular Shape Registration}
In Table~\ref{tables3}, we show the detailed CD, EMD, P2F for LV-endo, LV-epi and RV shape registration accuracy evaluations. Note that we register shapes from ED phase to ES phase. Our method outperforms all the baseline methods for LV-endo, LV-epi and RV shape registration, respectively.

\begin{table*}[t]
\begin{center}
\resizebox{2.0\columnwidth}{!}{
\begin{tabular}{l|c|c|c|c|c|c|c|c|c|c|c|c}
\hline
\multirow{2}{*}{Method} &\multicolumn{4}{c}{CD ($mm$) $\downarrow$} &\multicolumn{4}{|c|}{EMD ($mm$) $\downarrow$} &\multicolumn{4}{|c}{P2F ($mm$) $\downarrow$}\\
\cline{2-13}
 & Endo & Epi& RV & avg & Endo & Epi& RV & avg & Endo & Epi& RV & avg\\
\hline\hline
MR-Net &$2.19\pm0.754$ &$2.82\pm0.789$&$2.33\pm$0.737&$2.44\pm0.678$&$4.23\pm1.09$&$5.37\pm1.01$&$4.63\pm1.11$&$4.74\pm0.635$&$1.21\pm0.560$&$1.44\pm0.595$&$1.35\pm0.627$&$1.33\pm0.532$\\
NMF &$3.39\pm2.48$ &$3.85\pm1.36$&$4.02\pm2.34$&$3.76\pm1.84$&$4.18\pm1.08$&$5.94\pm1.25$&$4.63\pm1.28$&$4.92\pm0.909$&$2.19\pm1.91$&$2.26\pm1.07$&$2.90\pm2.19$&$2.45\pm1.54$\\
Ours-un &$1.99\pm0.635$ &$3.58\pm0.682$&$3.51\pm1.58$&$3.03\pm0.739$&$3.75\pm0.715$&$5.37\pm0.803$&$4.12\pm1.05$&$4.41\pm0.545$&$1.21\pm0.484$&$2.15\pm0.486$&$2.37\pm1.42$&$1.91\pm0.603$\\
Ours &$\mathbf{1.54}\pm0.535$ &$\mathbf{2.36}\pm0.584$&$\mathbf{1.91}\pm0.714$&$\mathbf{1.94}\pm0.525$&$\mathbf{3.43}\pm0.653$&$\mathbf{5.18}\pm0.820$&$\mathbf{3.23}\pm0.872$&$\mathbf{3.95}\pm0.514$&$\mathbf{0.827}\pm0.389$&$\mathbf{1.15}\pm0.390$&$\mathbf{1.03}\pm0.621$&$\mathbf{1.00}\pm0.402$\\

\hline
\end{tabular}}
\end{center}
\caption{Detailed Chamfer distance (CD), earth mover’s distance (EMD) and point-to-surface distance (P2F) for shape registration accuracy evaluation of LV-endo, LV-epi and RV, respectively. We register shapes from ED phase to ES phase.}
\label{tables3}
\end{table*}

\end{document}